\documentclass[final,5p,times,twocolumn]{elsarticle}

\usepackage{graphicx}
\usepackage{algorithm}
\usepackage{algorithmic}
\usepackage{array}
\usepackage{amsmath}

\journal{ArXiv}

\begin{document}
\begin{frontmatter}
\title{Deep Learning with Eigenvalue Decay Regularizer}
\author{Oswaldo Ludwig}

\address{Department of Computer Science, KU Leuven, Belgium, (email: oswaldoludwig@gmail.com)}
%\maketitle
\begin{abstract}
This paper extends our previous work on regularization of neural networks using Eigenvalue Decay by employing a soft approximation of the dominant eigenvalue in order to enable the calculation of its derivatives in relation to the synaptic weights, and therefore the application of back-propagation, which is a primary demand for deep learning. Moreover, we extend our previous theoretical analysis to deep neural networks and multiclass classification problems. Our method is implemented as an additional regularizer in Keras, a modular neural networks library written in Python, and evaluated in the benchmark data sets Reuters Newswire Topics Classification, IMDB database for binary sentiment classification, MNIST database of handwritten digits and  CIFAR-10 data set for image classification.
\end{abstract}
\end{frontmatter}

\section{Introduction}
\label{intro}

One of the problems in Machine Learning is termed \textit{overfitting}. The error on the training data set is driven toward a small value; however, the error is large when new data are presented to the trained algorithm. This occurs because the algorithm does not learn to generalize when new situations are presented. This phenomenon is related to the models' complexity, in Vapnik sense, and can be minimized by using regularization techniques \cite{Foresee} and \cite{MacKay}.

Current deep learning models present good generalization capacity, despite having very high VC dimensions \cite{koiran}. This is mostly because of recent advances in regularization techniques, which control the size of the hypothesis space \cite{poggio2003mathematics}. Existing libraries for deep learning allow users to set constraints on network parameters and to apply penalties on parameters \cite{Yaochu} or activity of the model layers. These penalties are usually incorporated into the loss function that the network optimizes on a per-layer basis and can be understood as soft constraints.

In our previous paper \cite{ludwig2014eigenvalue} we proposed and analyzed a regularization technique named Eigenvalue Decay, aiming at improving the classification margin, which is an effective strategy to decrease the classifier complexity, in Vapnik sense, by taking advantage on geometric properties of the training examples within the feature space. However, our previous approach requires a highly computational demanding training method based on Genetic Algorithms, which is not suitable for deep learning. In this paper we utilize a soft approximation of the dominant eigenvalue, in order to enable the calculation of its derivatives in relation to the synaptic weights, aiming at the application of back-propagation. Moreover, we extend our previous theoretical analysis to deep neural networks and multiclass classification problems.

The paper is organized as follows: Section \ref{bib} briefly reports the state-of-the-art in neural network regularization, while Section \ref{Theory} defines the problem of training with Eigenvalue Decay and analyzes the relationship between such regularization method and the classification margin. In Section \ref{training} we explain how we implement this method in Keras. Section \ref{sec_experiments} reports the experiments, while Section \ref{conclusion} summarizes some conclusions.

\section{State-of-the-art}
\label{bib}

There are many regularization strategies available to the deep learning practitioner, most of them based on regularizing estimators, i.e. trading increased bias for reduced variance \cite{geman}. In this section we briefly describe some of the most usual regularization strategies, such as constraining the parameter values of the model, adding extra terms in the objective function to penalize overly high values of the parameters and a recently developed technique, inspired in ensemble methods, which combines multiple hypotheses that explain the training data.

$L^2$ weight decay is the most usual weight regularizer, and was theoretically analyzed in \cite{bartlett}, which concludes that the bounds on the expected risk of a multilayer perceptron (MLP) depends on the magnitude of the parameters rather than the number of parameters. In the work \cite{bartlett} the author showed that the misclassification probability can be bounded in terms of the empirical risk, the number of training examples, and a scale-sensitive version of the VC-dimension, known as the fat-shattering dimension\footnote{See Theorem 2 of \cite{bartlett}}, which can be upper-bounded in terms of the magnitudes of the network parameters, independently from the number of parameters\footnote{See Theorem 13 of \cite{bartlett}}. In short, as regards $L^2$ weight-decay, the work \cite{bartlett} only shows that such a method can be applied to control the capacity of the classifier space. However, the best known way to minimize the capacity of the classifier space without damaging the accuracy on the training data is to maximize the classification margin, which is the SVM principle. Unfortunately, from the best of our knowledge, there is no formal proof that weight decay can maximize the margin. Therefore, we propose the Eigenvalue Decay, for which it is possible to establish a relationship between the eigenvalue minimization and the classification margin.

Another commonly used weight regularizer is the $L^1$ weight decay, which results in a sparse solution, in the sense that some parameters have an optimal value near zero. The sparsity property induced by $L^1$ regularization is useful as a feature selection mechanism, such as in the least absolute shrinkage and selection operator (LASSO) algorithm \cite{tibshirani1996regression}, which integrates an $L^1$ penalty with a linear model to lead a subset of the model weights to become zero.

It is also possible to set constraints on network parameters (usually on the norm of the parameters) during optimization, yielding a constrained optimization problem. If the $L^2$ norm is adopted, the weights are constrained to lie in an ball, resulting a smaller hypothesis space.

Early stopping \cite{treadgold} is a commonly employed method to improve the generalization capacity of neural networks (NN). This method also acts as a regularizer, since it restricts the optimization procedure to a small volume of parameter space within the neighborhood of the initial parameter value \cite{bishop95regularization}. In early stopping, the labeled data are divided into training and validation data sets. After some number of iterations the NN begins to overfit the data and the error on the validation data set begins to rise. When the validation error increases during a specified number of iterations, the algorithm stops the training section and applies the weights and biases at the minimum of the validation error to the NN.

The recently proposed DropOut \cite{srivastava2014dropout} provides a powerful way of regularizing deep models, while maintaining a relatively small computational cost. DropOut can be understood as a practical technique for constructing bootstrap aggregating (bagging) ensembles \cite{Ludwig_HOG} of many large NNs, i.e. DropOut trains an ensemble consisting of all sub-NNs that can be formed by removing non-output units from an underlying base NN \cite{Goodfellow-et-al-2016-Book}. However, while in bagging the models are all independent, in DropOut the models share parameters from the parent NN, making it possible to represent an exponential number of models with a tractable amount of memory.

\section{Eigenvalue Decay for deep neural networks}
\label{Theory}

In this section we define the problem of using Eigenvalue Decay in deep learning and show a relationship between this regularizer and the classification margin. 

Eigenvalue Decay can be understood as a weight decay regularizer; however, while the usual weight decay regularizers penalize overly high values of weights, Eigenvalue Decay penalizes overly high values of the dominant eigenvalue of $W_kW_k^T$, where $W_k$ is the synaptic weight matrix of any arbitrary layer $k$ of the NN. Both methods force the NN response to be smoother and less likely to overfit the training data.

We consider a binary or multiclass classification problem where the target output is encoded in one-hot style. Therefore, assuming $L$ classes, the target output is a $L$-dimensional vector where the position corresponding to the target class has the value $1$ and all the other $L-1$ positions have the value $-1$.

We analyze the classification margin of a deep MLP with $K$ hidden layer and linear output layer, whose model is given by:
\begin{equation}
\label{modela}
\begin{array}{l}
			y_{h_1}=\sigma\left(W_{1\left. \right.} x+b_{1}\right) \\ 
			\vdots \\
      y_{h_K}=\sigma\left(W_{K\left. \right.} y_{h_{K-1}}+b_{K}\right) \\       
      \hat{y}=W_{K+1 \left. \right.}y_{h_K}+b_{K+1} 
\end{array}
\end{equation}
where $y_{h_k}$ is the output vector of the $k^{th}$ hidden layer, $W_{k}$ is a matrix whose elements are the synaptic weights of layer $k$, $b_{k}$ is the bias vector of the layer $k$, $x$ is the input vector, and $\sigma\left(\cdot\right)$ is a commonly used activation function, such as the sigmoid function.

A single multiclass problem can be reduced into multiple binary classification problems; therefore, a multiclass classifier can be understood as an ensemble of binary classifiers which distinguish between one of the labels and the remainder (i.e. one-versus-all approach). In our case, a binary classifier for any arbitrary class $l$ can be built by substituting the matrix $W_{K+1}$ in (\ref{modela}) by its line $l$. Our analysis considers the classification margin per class, i.e. we analyze the relation between the application of Eigenvalue Decay on any hidden layer and the classification margin of the model (\ref{modela}) for any arbitrary class $l$, i.e. by considering only the line $l$ of the matrix $W_{K+1}$ in our analysis.

The MLP training using Eigenvalue Decay is modeled as:
\begin{equation}
\label{MSE_minlamb}
\min_{W_1,\ldots, W_{K+1},b_1,\ldots, b_{K+1}} E
\end{equation}
where
\begin{equation}
\label{E}
E=e+ \sum_{k=1}^{K}C_k\sqrt{\lambda^{dom}_k}\\
\end{equation}
where $e$ can be any commonly used loss function, such as mean squared error (MSE) or the Hinge loss, $\lambda^{dom}_k$ is the dominant eigenvalue of $W_{k}W_{k}^{T}$, $C_k$ is a constant that controls the regularization on the layer $k$ and $W_1,\ldots, W_{K+1}$, $b_1,\ldots, b_{K+1}$ are the weight matrices of the MLP model with $K+1$ layers, as defined in (\ref{modela}).

As can be seen in (\ref{MSE_minlamb}), our new theoretical analysis assumes the regularization by Eigenvalue Decay for any layer $l\in{1,\ldots,K}$, since our code enables the application of this regularizer on any hidden layer. We extended our previous analysis \cite{ludwig2014eigenvalue} by applying the chain rule to calculate the derivatives of the estimated output of the MLP in relation to the synaptic weights of any hidden layer, thus yielding larger equations and also unavoidable complexity. So, we recommend the reading of the theoretical derivations in \cite{ludwig2014eigenvalue} for MLP with only one hidden neuron before the reading of the derivations presented in the present paper.

We start our analysis with the following lemma: 
\begin{flushleft}
\end{flushleft}
\textbf{Lemma 1.} \cite{horn} \textit{ Let $\mathcal{K}$ denote the field of real numbers, $\mathcal{K}^{n\times n}$ a vector space containing all matrices with $n$ rows and $n$ columns with entries in $\mathcal{K}$, $A \in \mathcal{K}^{n\times n}$ be a symmetric positive-semidefinite matrix and $\lambda^{dom}$ be the dominant eigenvalue of $A$. Therefore, for any $x\in \mathcal{K}^{n}$, the following inequality holds true:}
\begin{equation}
\label{margin_product}
x^{T}Ax \leq \lambda^{dom} x^{T}x
\end{equation}
\begin{flushleft}
\end{flushleft}
\noindent

Our method penalizes the dominant eigenvalue aiming at maximizing the lower bound of the classification margin, as will be shown in Theorem 1. For the sake of space, we call the $l^{th}$ line of $W_{K+1}$ as $w_l$.

\begin{figure}[ht]
\vskip 0.0in
\begin{center}
\centerline{\includegraphics[width=\columnwidth]{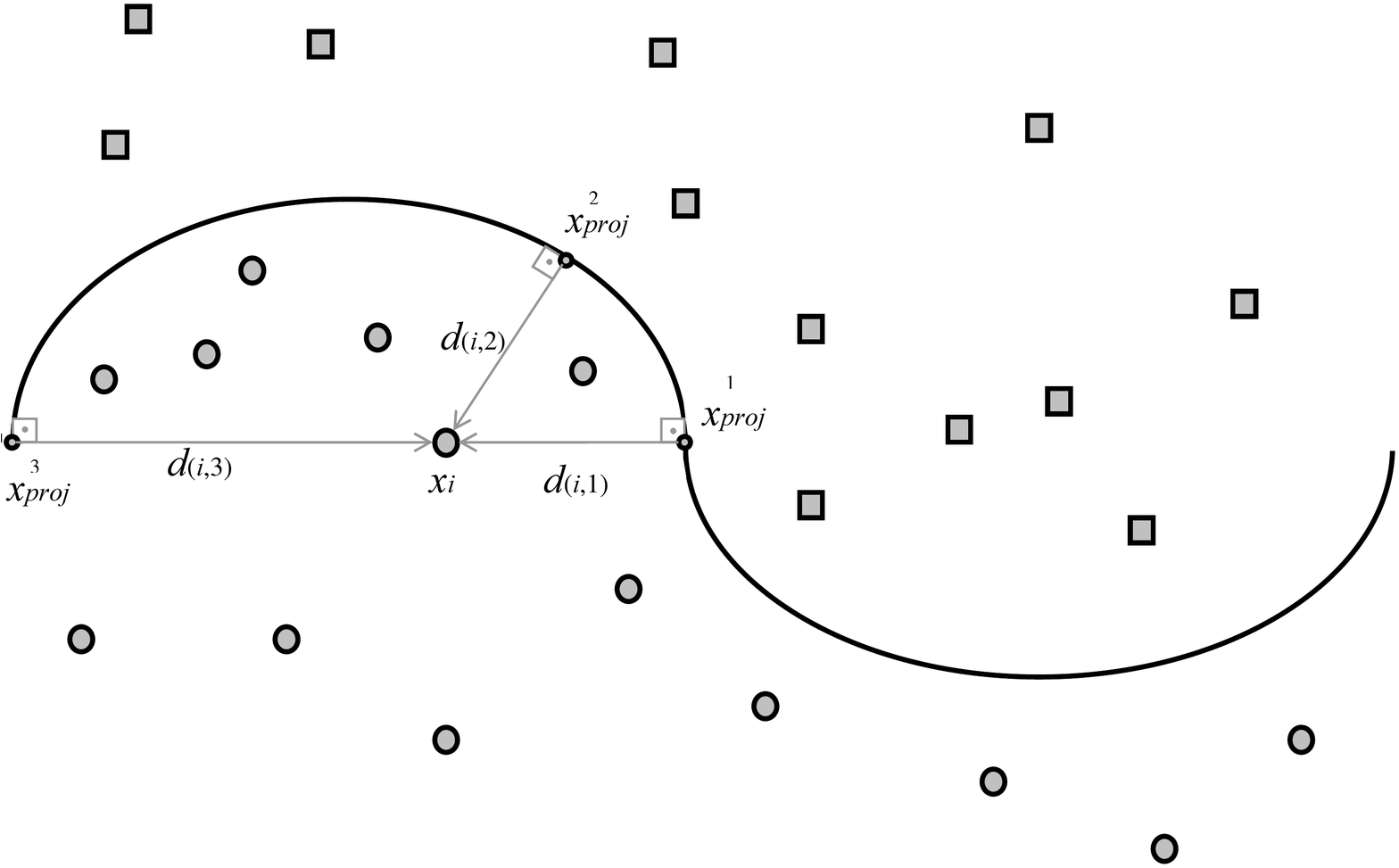}}
\caption{A feature space representing a separating surface with examples of the projections, $x^{j}_{proj}$, of the input $x_i$ and examples of orthogonal distances $d_{\left(i,j\right)}$.}
\label{fig_margin}
\end{center}
\vskip -0.3in
\end{figure}

We define the classification margin of the input data $x_i$ as the smallest orthogonal distance, $d_{\left(i,j\right)}$, between $x_i$ and the separating hypersurface defined by the MLP, see Figure \ref{fig_margin}.

\begin{flushleft}
\end{flushleft}

\begin{flushleft}
\end{flushleft}
\textbf{Theorem 1.} \textit{ Let $m_i$ be the classification margin of the training example $x_i$, for any arbitrary class $l$, and $\lambda^{dom}_k$ be the dominant eigenvalue of $W_{k}W_{k}^{T}$; then, for $m_{i}>0$, i.e. an example correctly classified, and a MLP with $K$ hidden layers, the following inequality hold true:}

\begin{equation}
\label{margin_theorem}
\frac{1}{\prod_{k=1}^K\sqrt{\lambda^{dom}_{k}}}\mu \leq m_i
\end{equation}
where
\begin{equation}
\label{margin_compos}
\mu=\min_{j}\left(\frac{y_{i}^{l}w_l\Omega^T\left(x_i-x_{proj}^{j}\right)}{\left\|w_l\right\|\sqrt{\prod_{k=1}^K\lambda^{dom}_{\left(activ,k\right)}}}\right),
\end{equation}
\noindent
\begin{equation}
\Omega=\prod_{n=1}^K\left(W_{n}^T\Gamma_{\left(n,j\right)}^T\right)
\end{equation}
\begin{equation}
\Gamma_{\left(k,j\right)}=\left.\frac{d^{ }y_{h_{k}}}{d^{ }v_k}\right|_{x_{proj}^j}
\end{equation}
\begin{equation}
v_k=W_{k\left. \right.} y_{h_{k-1}}+b_{k},
\end{equation}
$x_{proj}^{j}$ is the $j^{th}$ projection of $x_i$ on the separating hypersurface defined by the MLP, as illustrated in Fig.\ref{fig_margin}, $y_{i}^{l}$ is the $i^{th}$ target output for the class $l$, i.e. the position $l$ of the target vector $y$ and $\lambda^{dom}_{\left(activ,k\right)}$ is the dominant eigenvalue of $\Gamma_{\left(k,j\right)}\Gamma_{\left(k,j\right)}^T$.
\begin{flushleft}
\end{flushleft}

\begin{flushleft}
\end{flushleft}
{\bf Proof}. The first step in this proof is the calculation of the gradient of the position $l$ of the estimated output vector, $\hat{y}^{l}$, in relation to the input $x$ at the projected point $x^{j}_{proj}$ (see Figure \ref{fig_margin}):
\begin{equation}
\label{grad}
\nabla \hat{y}_{\left(i,j\right)}^{l}=\left.\frac{d \hat{y}^{l}}{d x}\right|_{x_{proj}^{j}}
\end{equation}
The normalized vector
\begin{equation}
\label{versor}
\vec{p}_{\left(i,j\right)}=\frac{\nabla \hat{y}_{\left(i,j\right)}^{l}}{\left\|\nabla \hat{y}_{\left(i,j\right)}^{l}\right\|}
\end{equation}
is normal to the separating surface, giving the direction from $x_i$ to $x_{proj}^{j}$; therefore
\begin{equation}
\label{distance}
x_i-x_{proj}^{j}= d_{\left(i,j\right)}\vec{p}_{\left(i,j\right)}
\end{equation}
where $d_{\left(i,j\right)}$ is the scalar distance between $x_i$ and $x_{proj}^{j}$. From (\ref{distance}) we have:
\begin{equation}
\label{distance2}
\nabla \hat{y}_{\left(i,j\right)}^{l}\left(x_i-x_{proj}^{j}\right)= d_{\left(i,j\right)}\nabla \hat{y}_{\left(i,j\right)}^{l}\vec{p}_{\left(i,j\right)}
\end{equation}
Substituting (\ref{versor}) into (\ref{distance2}) and solving for $d_{\left(i,j\right)}$, yields:
\begin{equation}
\label{margin}
d_{\left(i,j\right)}=\frac{\nabla \hat{y}_{\left(i,j\right)}^{l}\left(x_i-x_{proj}^{j}\right)}{\left\|\nabla \hat{y}_{\left(i,j\right)}^{l}\right\|}
\end{equation}
The sign of $d_{\left(i,j\right)}$ depends on which side of the decision surface $x_i$ is placed. It means that an example, $x_{i}$, correctly classified whose target value for the class $l$ is $-1$ corresponds to $d_{\left(i,j\right)}<0$. On the other hand, the classification margin must be positive in cases where examples are correctly classified, and negative in cases of misclassified examples, independently from their target classes. Therefore, the margin is defined as function of $y_{i}^{l}d_{\left(i,j\right)}$, where $y_{i}^{l}\in\left\{-1,1\right\}$ is the value of the target output of the $i^{th}$ training example for the class $l$. More specifically, the margin, $m_i$, is the smallest value of $y_{i}^{l}d_{\left(i,j\right)}$ in relation to $j$, that is:
\begin{equation}
\label{margin24}
m_{i}=\min_{j}\left(y_{i}^{l}d_{\left(i,j\right)}\right)
\end{equation}
Substituting (\ref{margin}) in (\ref{margin24}) yields:
\begin{equation}
\label{margin25}
m_i=\min_{j}\left(y_{i}^{l}\frac{\nabla \hat{y}_{\left(i,j\right)}^{l}\left(x_{i}-x_{proj}^{j}\right)}{\left\|\nabla \hat{y}_{\left(i,j\right)}^{l}\right\|}\right)
\end{equation}
For a MLP with a single hidden layer we have:
\begin{equation}
\label{jacobK}
\nabla \hat{y}_{\left(i,j\right)}^{l}=\left.\frac{d \hat{y}^{l}}{d x}\right|_{x_{proj}^{j}}=w_l\Gamma_{\left(1,j\right)} W_{1}
\end{equation} 
as can be derived from (\ref{modela}). Substituting (\ref{jacobK}) in (\ref{margin25}), yields:
\begin{equation}
\label{margin3}
m_i=\min_{j}\left(y_{i}^{l}\frac{w_l\Gamma_{\left(1,j\right)} W_{1}\left(x_i-x_{proj}^{j}\right)}{\sqrt{w_l\Gamma_{\left(1,j\right)} W_{1}W_{1}^{T}\Gamma_{\left(1,j\right)}^{T}w_l^{T}}}\right)
\end{equation}
Note that $W_{1}W_{1}^{T}$ is a symmetric positive-semidefinite matrix, therefore, from Lemma 1, the inequality:
\begin{equation}
\label{matrixnorm}
w_l\Gamma_{\left(1,j\right)} W_{1}W_{1}^{T}\Gamma_{\left(1,j\right)}^{T}w_l^{T}\leq \lambda^{dom}_K w_l\Gamma_{\left(1,j\right)} \Gamma_{\left(1,j\right)}^{T}w_l^{T}
\end {equation}
holds true for any $\Gamma_{\left(K,j\right)}$ and any $w_l$. Therefore, we can write:
\begin{equation}
\label{margin40}
m_i\geq \frac{1}{\sqrt{\lambda^{dom^{ }}_1}} \min_{j}\left(y_{i}^{l}\frac{w_l\Gamma_{\left(1,j\right)} W_{1}\left(x_i-x_{proj}^{j}\right)}{\sqrt{w_l\Gamma_{\left(1,j\right)}\Gamma_{\left(1,j\right)}^{T}w_l^{T}}}\right)
\end{equation}
Since $\Gamma_{\left(1,j\right)}\Gamma_{\left(1,j\right)}^{T}$ is also a symmetric positive-semidefinite matrix:
\begin{equation}
\label{margin4}
m_i\geq \frac{1}{\sqrt{\lambda^{dom^{ }}_1}} \min_{j}\left(y_{i}^{l}\frac{w_l\Gamma_{\left(1,j\right)} W_{1}\left(x_i-x_{proj}^{j}\right)}{\left\|w_l\right\|\sqrt{ {\lambda^{dom^{ }}_{\left(activ,1\right)}}}}\right)
\end{equation}
For a MLP with two hidden layers we have:
\begin{equation}
\label{jacobK2}
\nabla \hat{y}_{\left(i,j\right)}^{l}=\left.\frac{d \hat{y}^{l}}{d x}\right|_{x_{proj}^{j}}=w_l\Gamma_{\left(2,j\right)} W_{2}\Gamma_{\left(1,j\right)} W_{1}
\end{equation} 
Substituting (\ref{jacobK2}) in (\ref{margin25}), yields:
\begin{equation}
\label{margin5}
%\begin{array}{l}
m_i=
\min_{j}\left(y_{i}^{l}\frac{w_l\Gamma_{\left(2,j\right)} W_{2}\Gamma_{\left(1,j\right)} W_{1}\left(x_i-x_{proj}^{j}\right)}{\sqrt{w_l\Gamma_{\left(2,j\right)} W_{2}\Gamma_{\left(1,j\right)} W_{1}W_{1}^T\Gamma_{\left(1,j\right)}^TW_{2}^{T}\Gamma_{\left(2,j\right)}^{T}w_l^{T}}}\right)
%\end{array}
\end{equation}
Since $W_{1}W_{1}^{T}$, $W_{2}W_{2}^{T}$, $\Gamma_{\left(1,j\right)}\Gamma_{\left(1,j\right)}^{T}$ and $\Gamma_{\left(2,j\right)}\Gamma_{\left(2,j\right)}^{T}$ are symmetric positive-semidefinite matrices:
\begin{equation}
\label{margin6}
%\begin{array}{l}
m_i\geq
\frac{1}{\sqrt{\lambda^{dom^{ }}_{1}\lambda^{dom^{ }}_{2}}} \min_{j}\left(y_{i}^{l}\frac{w_l\Gamma_{\left(2,j\right)} W_{2}\Gamma_{\left(1,j\right)} W_{1}\left(x_i-x_{proj}^{j}\right)}{\left\|w_l\right\|\sqrt{\lambda^{dom^{ }}_{\left(activ,1\right)}\lambda^{dom^{ }}_{\left(activ,2\right)}}}\right)
%\end{array}
\end{equation}
From (\ref{margin4}) and (\ref{margin6}) we can deduce (\ref{margin_theorem}) by induction.% $\blacksquare$

Taking into account that $\sqrt{\lambda^{dom}_k}$ is in the denominator of the bound in (\ref{margin_theorem}), the training method based on Eigenvalue Decay decreases $\sqrt{\lambda^{dom}_k}$ aiming at increasing the lower bound on the classification margin. However, Eigenvalue Decay does not assure, by itself, increasing the margin, because $\mu_k$ is function of $W_{k}$.
 
\section{Using Eigenvalue Decay in Keras}
\label{training}

The use of Eigenvalue Decay within a deep learning library, such as Keras \cite{chollet2015}, requires, not only a lightweight algorithm, but also a formulation that enables the calculation of derivatives of the objective function in relation to the synaptic weights, aiming at the application of back-propagation, which is a main demand for deep learning.

We approximate the dominant eigenvalue by the power method; therefore, assuming $M=W_{k}W_{k}^{T}$, the eigenvector corresponding to the dominant eigenvalue can be approximated by:
\begin{equation}
\label{power}
v^{dom}=M^{^{p}}v_1
\end{equation}
where $p$ is an arbitrary positive integer and $v_1$ is an initial nonzero approximation of the dominant eigenvector. We set all the elements of $v_1$ equal to one. Having $v^{dom}$, we calculate $\lambda^{dom}$ as follows:
\begin{equation}
\label{eigenapprox}
\lambda^{dom}=\frac{Mv^{dom}\cdot v^{dom}}{v^{dom} \cdot v^{dom}}
\end{equation}

The approximation given by the power method has derivatives in relation to the synaptic weights, i.e. the elements of $W_{k}$, enabling the application of backpropagation in Keras. We implement Eigenvalue Decay in Keras using Theano functions to model the approximation of $\lambda^{dom}$ based on an approximation of $v^{dom}$ where $p=9$. Our source
code is freely available in Github\footnote{https://github.com/oswaldoludwig/Eigenvalue-Decay-Regularizer-for-Keras}.

Beyond the custom regularizer presented in this paper, it is possible to implement a custom objective function\footnote{https://github.com/oswaldoludwig/visually-informed-embedding-of-word-VIEW-} in Keras, see our previous work \cite{ludwig2016deep}, where we implemented a custom version of the Hinge loss, aiming at a SVM-like learning for multiclass classification. The idea is to make better use of the margin resulting from the use of Eigenvalue Decay, since the Hinge loss penalizes only examples that violate a given margin or are misclassified, i.e. an estimated output smaller than 1 in response to a positive example or an estimated output larger than -1 in response to a negative example (these training examples can be understood as support vectors). The other training examples are ignored during the optimization, i.e. they don't participate in defining the decision surface.

%-----------------------------------------------------------------------------------
\section {Experiments}
\label{sec_experiments}
In this section our methods are evaluated using the benchmark data sets Reuters Newswire Topics Classification, IMDB database for binary sentiment classification, MNIST database of handwritten digits and CIFAR-10 data set for image classification.

Reuters Newswire Topics Classification (RNTC) is a collection of 11228 newswires from Reuters, labeled over 46 topics. IMDB Movie Reviews Sentiment Classification Dataset is a collection of 25000 movies reviews from IMDB, labeled by sentiment (positive/negative). MNIST database of Handwritten Digits Dataset is a collection of 60000 $28\times28$ grayscale images of the 10 digits, along with a test set of 10000 images. CIFAR-10 data set consists of 60000 $32\times32$ color images labeled over 10 categories, 50000 training images and 10000 test images. Table \ref{datasets} summarizes the details of the data sets.

\begin{table}[!hbt]
\renewcommand{\arraystretch}{1.3}
\caption{Datasets used in the experiments}
\label{datasets}
\centering
\footnotesize
%\scriptsize
%\tiny
\begin{tabular}{lccc}
\hline
Dataset & attributes & \# data train & \# data test\\
\hline\hline
RNTC & 1000 & 8982 & 2246 \\
IMDB & 100 & 20000 & 5000 \\
MNIST & 784 & 60000 & 10000 \\
CIFAR-10 & $32\times32\times3$ & 50000 & 10000 \\
\hline
\end{tabular}
\end{table}

For the sake of comparison, we adopt the original models from the examples provided in the Keras repository \footnote{https://github.com/fchollet/keras/tree/master/examples}, which are well adjusted and regularized with DropOut. Our experiments compare the performance of weight regularizers; therefore, we keep the original DropOut regularization and apply Eigenvalue Decay (ED), $L^1$ and $L^2$ weight regularizers to compare the accuracy gains over the original models from Keras repository.

Among the models available in the Keras repository, we adopt the MLP with a single hidden layer for RNTC, the deep pipeline composed by an embedding layer, a convolutional neural network (CNN) \cite{lecun1995convolutional}, a long short term memory (LSTM) network \cite{hochreiter1997long} and a dense layer for IMDB, the deep MLP with two hidden layers for MNIST and the deep pipeline composed by a MLP stacked on the top of a CNN for CIFAR-10. The experiment with CIFAR-10 does not use data augmentation. In our experiments we apply the weight regularizers in both layers of the MLP used for RNTC, on the dense and embedding layers of the deep model used for IMBD, on the last two layers of the deep MLP used for MNIST and on both dense layers of the model used for CIFAR-10.

To find the optimal values of $C_k$ in (\ref{E}), we exploit a 2D grid using 5-fold cross validation on the training data, keeping the original architectures and loss functions of the Keras models, i.e. categorical cross-entropy for RNTC, MNIST and CIFAR-10, and binary cross-entropy for IMDB. The accuracy values on the test data and the processing time per training epoch, running in a GPU NVIDIA GeForce GTX 980, are summarized in Table \ref{gains}, where $\Delta$ is the gain over the original model from Keras. In the case of the IMDB and CIFAR-10 data sets the accuracy values were averaged over 10 runs.
\renewcommand{\arraystretch}{1}
\begin{table}[!hbt]
\caption{Accuracy ($acc$), gain over the original model from Keras ($\Delta$) and processing time (in seconds) for the original model regularized with DropOut and using DropOut together with each of the three weight regularizers.}
\label{gains}
\centering
\footnotesize
\begin{tabular}{l c c c c}
\hline
& DropOut & DropOut+ED & DropOut+$L^1$ & DropOut+$L^2$\\
\hline\hline
\multicolumn{5}{c}{RNTC} \\
\hline
$acc (\%)$ & 79.78 & 80.72 & 79.78 & 80.45\\
$\Delta (\%)$  & 0.00 & 0.94 & 0.00 & 0.67 \\
$sec/epoch$ & $<1$ & 5 & $<1$ & $<1$\\
\hline
\multicolumn{5}{c}{IMDB} \\
\hline
$acc (\%)$ & 84.98 & 85.36 & 85.08 & 85.05 \\
$\Delta (\%)$    & 0.00 & 0.38 & 0.10 & 0.07 \\
$sec/epoch$ & 17 & 23 & 19 & 19\\
\hline
\multicolumn{5}{c}{MNIST} \\
\hline
$acc (\%)$ & 98.40 & 98.78 & 98.57 & 98.68\\
$\Delta (\%)$ & 0.00 & 0.38 & 0.17 & 0.28\\
$sec/epoch$ & 2 & 9 & 2 & 2\\
\hline
\multicolumn{5}{c}{CIFAR-10} \\
\hline
$acc (\%)$ & 77.95 & 80.66 & 79.31 & 79.11\\
$\Delta (\%)$ & 0.00 & 2.71 & 1.36 & 1.16\\
$sec/epoch$ & 39 & 67 & 41 & 41\\
\hline
\end{tabular}
\end{table}

As can be seen in Table \ref{gains}, the weight regularizers yielded small gains on the accuracy, since the models provided in the Keras repository are well adjusted and regularized with DropOut. Eigenvalue Decay yielded the largest gains in all the data sets, but it was also the most costly regularizer, which is not a surprise, given the cost associated with the computation of the dominant eigenvalue by the power method.

\section{Conclusion}
\label{conclusion}

This work introduces a new option of weight regularizer to the deep learning practitioners. The analysis presented in this paper indicates that Eigenvalue Decay can increase the classification margin, which can improve the generalization capability of deep models.

In the scope of weight regularizers, the experiments indicate that Eigenvalue Decay can provide better gains on the classification accuracy at the cost of a larger CPU/GPU time.

%\bibliographystyle{unsrt}

%\begin{thebibliography}{99}

%\bibliographystyle{model1a-num-names}
\bibliographystyle{elsarticle-num}

\bibliography{EigenvalueDecayKeras_ArXiv_v3}

\begin{thebibliography}{10}
\expandafter\ifx\csname url\endcsname\relax
  \def\url#1{\texttt{#1}}\fi
\expandafter\ifx\csname urlprefix\endcsname\relax\def\urlprefix{URL }\fi
\expandafter\ifx\csname href\endcsname\relax
  \def\href#1#2{#2} \def\path#1{#1}\fi

\bibitem{Foresee}
F.~Dan~Foresee, M.~Hagan, Gauss-newton approximation to bayesian learning, in:
  Neural Networks, 1997., International Conference on, Vol.~3, IEEE, 1997, pp.
  1930--1935.

\bibitem{MacKay}
D.~MacKay, Bayesian interpolation, Neural computation 4~(3) (1992) 415--447.

\bibitem{koiran}
P.~Koiran, E.~Sontag, Neural networks with quadratic {VC} dimension, Journal of
  Computer and System Sciences 54~(1) (1997) 190--198.

\bibitem{poggio2003mathematics}
T.~Poggio, S.~Smale, The mathematics of learning: Dealing with data, Notices of
  the AMS 50~(5) (2003) 537--544.

\bibitem{Yaochu}
Y.~Jin, Neural network regularization and ensembling using multi-objective
  evolutionary algorithms, in: In: Congress on Evolutionary Computation
  (CEC'04), IEEE, IEEE Press, 2004, pp. 1--8.

\bibitem{ludwig2014eigenvalue}
O.~Ludwig, U.~Nunes, R.~Araujo, Eigenvalue decay: A new method for neural
  network regularization, Neurocomputing 124 (2014) 33--42.

\bibitem{geman}
S.~Geman, E.~Bienenstock, R.~Doursat, Neural networks and the bias/variance
  dilemma, Neural computation 4~(1) (1992) 1--58.

\bibitem{bartlett}
P.~Bartlett, The sample complexity of pattern classification with neural
  networks: the size of the weights is more important than the size of the
  network, Information Theory, IEEE Transactions on 44~(2) (1998) 525--536.

\bibitem{tibshirani1996regression}
R.~Tibshirani, Regression shrinkage and selection via the lasso, Journal of the
  Royal Statistical Society. Series B (Methodological) (1996) 267--288.

\bibitem{treadgold}
N.~Treadgold, T.~Gedeon, Exploring constructive cascade networks, Neural
  Networks, IEEE Transactions on 10~(6) (1999) 1335 --1350.
\newblock \href {http://dx.doi.org/10.1109/72.809079}
  {\path{doi:10.1109/72.809079}}.

\bibitem{bishop95regularization}
C.~Bishop, Regularization and complexity control in feed-forward networks, in:
  Proceedings International Conference on Artificial Neural Networks ICANN'95,
  Vol.~1, EC2 et Cie, pp. 141--148.

\bibitem{srivastava2014dropout}
N.~Srivastava, G.~Hinton, A.~Krizhevsky, I.~Sutskever, R.~Salakhutdinov,
  Dropout: A simple way to prevent neural networks from overfitting, The
  Journal of Machine Learning Research 15~(1) (2014) 1929--1958.

\bibitem{Ludwig_HOG}
O.~Ludwig, D.~Delgado, V.~Goncalves, U.~Nunes, Trainable classifier-fusion
  schemes: An application to pedestrian detection, in: In Proceedings of the
  12th International {IEEE} Conference on Intelligent Transportation Systems,
  {ITSC}2009, 2009, pp. 1--6.

\bibitem{Goodfellow-et-al-2016-Book}
Y.~B. Ian~Goodfellow, A.~Courville, \href{http://www.deeplearningbook.org}{Deep
  learning}, book in preparation for MIT Press (2016).
\newline\urlprefix\url{http://www.deeplearningbook.org}

\bibitem{horn}
R.~Horn, C.~Johnson, Matrix analysis, Cambridge university press, 1990.

\bibitem{chollet2015}
F.~Chollet, Keras: Theano-based deep learning library, Code: https://github.
  com/fchollet. Documentation: http://keras. io.

\bibitem{ludwig2016deep}
O.~Ludwig, X.~Liu, P.~Kordjamshidi, M.-F. Moens, Deep embedding for spatial
  role labeling, arXiv preprint arXiv:1603.08474.

\bibitem{lecun1995convolutional}
Y.~LeCun, Y.~Bengio, Convolutional networks for images, speech, and time
  series, The handbook of brain theory and neural networks 3361~(10) (1995)
  1995.

\bibitem{hochreiter1997long}
S.~Hochreiter, J.~Schmidhuber, Long short-term memory, Neural computation 9~(8)
  (1997) 1735--1780.

\end{thebibliography}

\end{document}